# A Comparative Study of Fingerprint Image-Quality Estimation Methods

Fernando Alonso-Fernandez, *Student Member, IEEE*, Julian Fierrez, *Member, IEEE*,
Javier Ortega-Garcia, *Member, IEEE*, Joaquin Gonzalez-Rodriguez, *Member, IEEE*, Hartwig Fronthaler,
Klaus Kollreider, and Josef Bigun, *Fellow, IEEE*

*Abstract*—One of the open issues in fingerprint verification is the lack of robustness against image-quality degradation. Poor-quality images result in spurious and missing features, thus degrading the performance of the overall system. Therefore, it is important for a fingerprint recognition system to estimate the quality and validity of the captured fingerprint images. In this work, we review existing approaches for fingerprint image-quality estimation, including the rationale behind the published measures and visual examples showing their behavior under different quality conditions. We have also tested a selection of fingerprint image–quality estimation algorithms. For the experiments, we employ the BioSec multimodal baseline corpus, which includes 19 200 fingerprint images from 200 individuals acquired in two sessions with three different sensors. The behavior of the selected quality measures is compared, showing high correlation between them in most cases. The effect of low-quality samples in the verification performance is also studied for a widely available minutiae-based fingerprint matching system.

## I. Introduction

BIOMETRIC authentication has been receiving considerable attention over the last years due to the increasing demand for automatic person recognition. The term "biometrics" refers here to automatic recognition of an individual based on behavioral and/or physiological characteristics (e.g., fingerprints, face, iris, voice, signature, etc.), which cannot be stolen, lost, or copied [1]. Among all biometric techniques, fingerprint recognition is the most widespread in personal identification systems due to its permanence and uniqueness [2]. Fingerprints are being increasingly used not only in forensic investigations, but also in a large number of convenience applications, such as access control or online identification [1].

This work was supported in part by BioSecure NoE, in part by the TIC2003-08382-C05-01 and TEC2006-13141-C03-03 projects of the Spanish Ministry of Science and Technology, in part by the Consejeria de Educacion de la Comunidad de Madrid and Fondo Social Europeo, and in part by the Marie Curie Fellowship from the European Commission.

F. Alonso-Fernandez, J. Fierrez, J. Ortega-Garcia, and J. Gonzalez-Rodriguez are with ATVS/Biometric Recognition Group, Politecnica Superior, University de Madrid, Madrid 28049, Spain (e-mail: fernando.alonso@uam.es; julian.fierrez@uam.es; javier.ortega@uam.es; joaquin.gonzalez@uam.es).

H. Fronthaler, K. Kollreider, and J. Bigun are with Halmstad University, Halmstad SE-30118, Sweden (e-mail: hartwig.fronthaler@ide.hh.se; klaus.kollreider@ide.hh.se; josef.bigun@ide.hh.se).

ISO/INCITS-M1[1] has recently established a biometric sample-quality draft standard [3], in which a biometric sample quality is considered from three different points of view: 1) character, which refers to the quality attributable to inherent physical features of the subject; 2) fidelity, which is the degree of similarity between a biometric sample and its source, attributable to each step through which the sample is processed; and 3) utility, which refers to the impact of the individual biometric sample on the overall performance of a biometric system, where the concept of sample quality is a scalar quantity that is related monotonically to the performance of the system [4]. The character of the sample source and the fidelity of the processed samples contributes to, or similarly detracts from, the utility of the sample. It is generally accepted that the utility is most importantly mirrored by a quality metric [4], [5], so that images assigned higher quality shall necessarily lead to better identification of individuals (i.e., better separation of genuine and impostor match score distributions). Some previous experiments of the utility of quality metrics are [6] and [7], in which the verification performance of fingerprint matchers is studied for different image-quality groups. The fidelity of quality metrics is studied in [8]–[10]. Wilson *et al.* [8] have studied the effects of image resolution in the matching accuracy, whereas Capelli *et al.* [9] have studied the correlation between the quality characteristics of a fingerprint scanner with the performance they can ensure when the acquired images are matched by a recognition algorithm. In [10], we can find a number of quality metrics aimed at objectively assessing the quality of an image in terms of the similarity between a reference image and a degraded version of it.

A theoretical framework for a biometric sample quality has been developed by Youmaran and Adler [5]. They relate biometric sample quality with the identifiable information contained. An approach to measure the loss of information due to quality degradation is proposed. "Biometric information" ($BI$) is defined in [5] as the decrease in uncertainty about the identity of a person due to a set of biometric measurements. $BI$ is calculated by the relative entropy between the population feature distribution and the person's feature distribution. Degradations to a biometric sample will reduce the amount of identifiable information. The results reported in [5] show that degraded biometric samples result in a decrease in $BI$.

A number of factors can affect the quality of fingerprint images [11]: occupation, motivation/collaboration of users, age,

---

[1]International Standards Organization/International Committee for Information Technology Standards.

temporal or permanent cuts, dryness/wetness conditions, temperature, dirt, residual prints on the sensor surface, etc. Unfortunately, many of these factors cannot be controlled and/or avoided. For this reason, assessing the quality of captured fingerprints is important for a fingerprint recognition system. There are many roles of a quality measure in the context of biometric systems [4]: 1) quality algorithms may be used as a monitoring tool [12]; 2) quality of enrolment templates and/or samples acquired during an access transaction can be controlled by acquiring until satisfaction (recapture); and 3) some of the steps of the recognition system can be adjusted based on the estimated quality (quality-based adaptation [13]).

Fingerprint quality is usually defined as a measure of the clarity of ridges and valleys and the extractability of the features used for identification such as minutiae, core and delta points, etc. [15]. In other words, most of the operational schemes for fingerprint image-quality estimation are focused on the utility of the images. In the rest of this paper, we follow this approach. A framework for evaluating and comparing quality measures in terms of their capability of predicting the system performance is presented in [4]. In this work, we follow this framework by reporting the equal error rate (EER), false acceptance rate (FAR), and false rejection rate (FRR) of the verification system as we reject samples with the lowest quality.

This paper presents a comprehensive survey of the fingerprint-quality algorithms found in the literature, extending a preliminary survey presented in [14]. We provide basic algorithmic descriptions of each quality estimation measure and the rationale behind them. We also include visual examples that show the behavior of the measures with fingerprint images of different quality. A selection of quality measures is also compared in terms of the correlation between them. In addition, to illustrate the importance of having a quality estimation step in fingerprint recognition systems, we study the effects of rejecting low-quality samples in the performance of a widely available fingerprint matching system that uses minutiae to represent and match fingerprints. We use for our experiments a real multisession and multisensor database [16]. To the best of our knowledge, no previous studies on the impact of image quality in fingerprint verification systems using a real multisession and multisensor database have been found in the literature.

The rest of this paper is organized as follows. We review existing algorithms for fingerprint image-quality estimation in Section II. The experiments, including a sketch of the fingerprint matcher used, the quality measures compared, the database, and the results are described in Section III. Conclusions are finally drawn in Section IV.

## II. ALGORITHMS FOR FINGERPRINT IMAGE-QUALITY ESTIMATION

Existing approaches for fingerprint image quality estimation can be divided into: 1) those that use local features of the image; 2) those that use global features of the image; and 3) those that address the problem of quality assessment as a classification problem. A summary of existing local and global fingerprint-quality measures, including a brief description, is shown in Tables I and II, respectively.

TABLE I
SUMMARY OF EXISTING FINGERPRINT-QUALITY
MEASURES BASED ON LOCAL FEATURES

| LOCAL DIRECTION |
|---|
| *Orientation Certainty Level* [17] |
| Orientation strength measure computed from the gradient of the gray level image |
| *Ridge frequency, ridge thickness, ridge-to-valley thickness* [17]. Computed from the sinusoid that models ridges and valleys in the direction normal to ridge flow |
| *Local Orientation* [18] |
| Average absolute difference of local orientation with the surrounding blocks |
| *Spatial Coherence* [15] |
| Direction coherence measure computed from the gradient of the gray level image |
| *Symmetry features* [19] |
| Correlation between linear and parabolic symmetry in a fingerprint image |
| **GABOR FILTERS** |
| *Gabor features* [20] |
| Standard deviation of $m$ filter responses with different directions |
| **PIXEL INTENSITY** |
| *Directionality* [21] |
| Minimum sum of intensity differences between a pixel $(i, j)$ and $l$ pixels selected along a line segment centered at $(i, j)$, computed for $n$ different directions of the line segment |
| *Variance and local contrast* [11] |
| *Mean, variation, contrast and eccentric moment* [22] |
| *Clustering Factor* [23] |
| Degree to which similar pixels (i.e. ridges or valleys) cluster in the nearby region |
| *Local Clarity* [18] |
| Overlapping area of the gray level distributions of segmented ridges and valleys |
| **POWER SPECTRUM** |
| *DFT of the sinusoid that models ridges and valleys* [23] |
| **COMBINATION OF LOCAL FEATURES** |
| *Amplitude, frequency and variance of the sinusoid that models ridges and valleys* [24] |
| *Direction map, low contrast map, low flow map and high curve map* [25] |

TABLE II
SUMMARY OF EXISTING FINGERPRINT-QUALITY
MEASURES BASED ON GLOBAL FEATURES

| DIRECTION FIELD |
|---|
| *Continuity of the direction field* [17] |
| Detection of abrupt direction changes between blocks |
| *Uniformity of the frequency field* [17] |
| Standard deviation of the ridge-to-valley thickness ratio |
| **POWER SPECTRUM** |
| *Energy concentration in ring-shaped regions of the spectrum* [15] |

### A. Methods Based on Local Features

Methods that rely on local features usually divide the image into nonoverlapped square blocks and extract features from each block. Blocks are then classified into groups of different quality. A local measure of quality is finally generated. This local measure can be the percentage of blocks classified with "high" or "low" quality, or an elaborated combination. Some methods assign a relative weight to each block based on its distance from the centroid of the fingerprint image, since blocks near the centroid are supposed to provide more reliable information [15], [21].

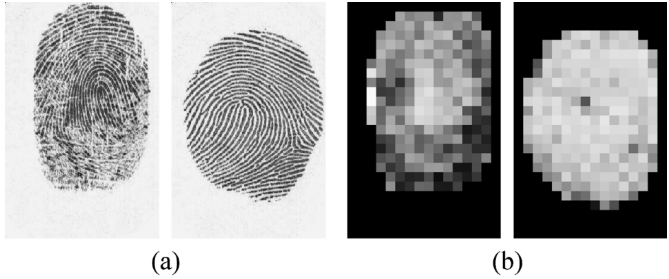

Fig. 1. Computation of the OCL for two fingerprints of different quality. Panel (a) is the input fingerprint images. Panel (b) is the blockwise values of the OCL; blocks with the brighter color indicate higher quality in the region.

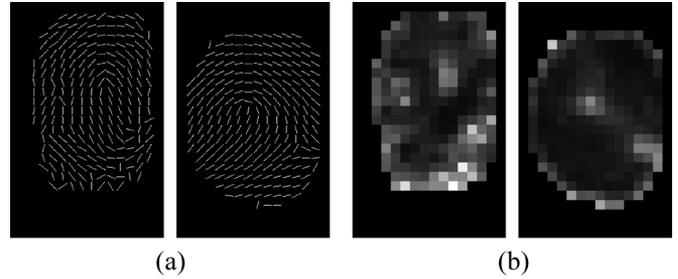

Fig. 2. Computation of the local orientation quality (LOQ) for two fingerprints of different quality. Panel (a) is the direction fields of the images shown in Fig. 1(a). Panel (b) is the blockwise values of the average absolute difference of local orientation with the surrounding blocks; blocks with a brighter color indicate higher difference value and, thus, lower quality.

*1) Based on the Local Direction:* This group of methods use the local direction information provided by the direction field [26] to compute several local features in each block. For a comprehensive introduction of the theory and applications of direction fields, we refer the reader to [27].

The method presented by Lim *et al.* [17] computes the following features in each block: Orientation certainty level (OCL), ridge frequency, ridge thickness and ridge-to-valley thickness ratio. Blocks are then labeled as "good," "undetermined," "bad," or "blank" by setting thresholds for the four features. A local quality score $S_L$ is finally computed based on the total number of "good," "undetermined," and "bad" quality image blocks in the image. The OCL measures the energy concentration along the dominant direction of ridges. It is computed as the ratio between the two eigenvalues of the covariance matrix of the gradient vector. Ridge frequency is used to detect abnormal ridges that are too close or too far whereas ridge thickness and ridge-to-valley thickness ratio are used to detect ridges that are unreasonably thick or thin. An example of OCL computation is shown in Fig. 1 for two fingerprints of different quality.

The OCL is also used in [23] to detect high curvature regions of the image. Although high curvature has no direct relationship with the quality of a fingerprint image (e.g., core and delta points), it could help to detect regions with invalid curvature. The curvature of a block is captured in [23] by combining the orientations of four quadrants and each of their certainty levels. Both measures are used together to distinguish between blocks with core/deltas and blocks with invalid curvature due to low quality.

The method presented in [18] computes the average absolute difference of local orientation with the surrounding blocks, resulting in a local orientation quality measure (LOQ). A global orientation quality score (GOQS) is finally computed by averaging all of the local orientation quality scores of the image. In high-quality images, it is expected that ridge direction changes smoothly across the whole image; thus, the GOQS provides information about how smoothly the local direction changes from block to block. An example of local orientation quality computation is shown in Fig. 2 for two fingerprints of different quality.

Recently, Chen *et al.* [15] proposed a local quality index which measures the local coherence of the intensity gradient, reflecting the clarity of local ridge-valley direction in each block. A local quality score $Q_S$ is finally computed by averaging the coherence of each block.

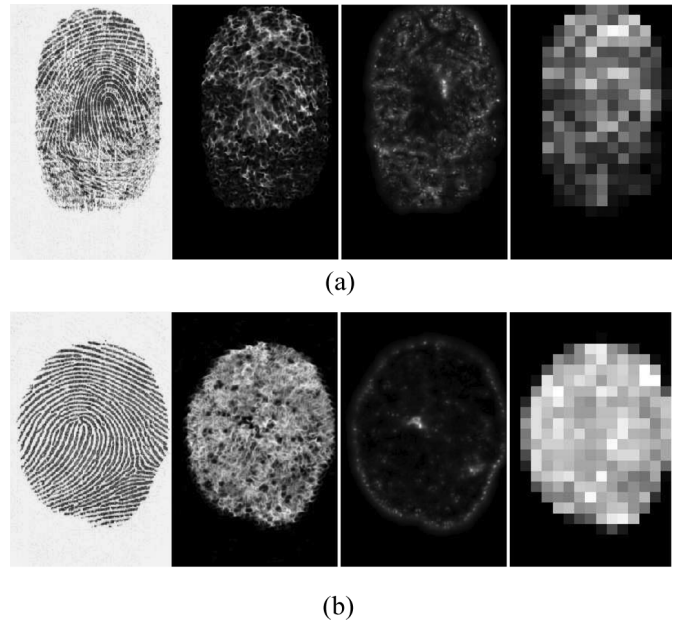

Fig. 3. Estimation of fingerprint quality using symmetry features. The figure shows the decomposition of two fingerprints of different quality into linear and parabolic symmetry (second and third column, respectively). The final local quality estimation in blocks is depicted in the fourth column (blocks with brighter color indicate higher quality in the region). (a) Low-quality fingerprint. (b) High-quality fingerprint.

The method presented in [19] employs symmetry features for fingerprint-quality assessment. In this approach, the orientation tensor [28] of a fingerprint image is decomposed into two symmetry representations, allowing to draw conclusions on its quality. On one hand, a coherent ridge flow has linear symmetry and is thus modeled by symmetry features of order 0. On the other hand, points of high curvature, such as minutia, core, and delta points exhibit parabolic symmetry and are therefore represented by symmetry features of order 1. Fig. 3 depicts these two symmetry representations for two fingerprints of different quality. In a further step, the two symmetries are combined and averaged within small nonoverlapped blocks, yielding $S_b$. To determine the final local quality $Q_b$, $S_b$ is negatively weighted with the blockwise correlation between the two involved symmetries. A large negative correlation is desirable in terms of quality, because this suggests well-separated symmetries. The local quality $Q_b$ is also visualized in the last column of Fig. 3.

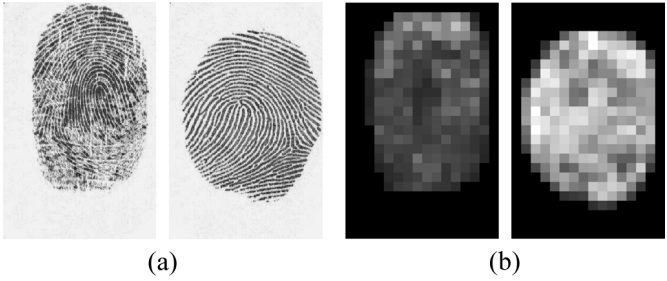

Fig. 4. Estimation of fingerprint quality using Gabor filters. Panel (a) is the input fingerprint images. Panel (b) is the blockwise values of the standard deviation of $m$ filter responses (eight in this example) with a different direction. Blocks with a brighter color indicate higher standard deviation value and, thus, higher quality.

An overall quality measure is derived by averaging over the foreground blocks of $Q_b$.

*2) Based on Gabor Filters:* Gabor filters can be viewed as a filter bank that can represent the local frequencies. Two-dimensional quadrature mirror filters are close akins of Gabor filters [29]. Gabor filters were introduced to image processing by [30], and both filter families represent another implementation of the local-direction fields [27], though they are frequently used stand alone, without a local-direction field interpretation.

Shen *et al.* [20] proposed a method based on Gabor features. Each block is filtered using a Gabor filter with $m$ different directions. If a block has high quality (i.e., strong ridge direction), one or several filter responses are larger than the others. In poor-quality blocks or background blocks, the $m$ filter responses are similar. The standard deviation of the $m$ filter responses is then used to determine the quality of each block ("good" and "poor"). A quality index $QI$ of the whole image is finally computed as the percentage of foreground blocks marked as "good." If $QI$ is lower than a predefined threshold, the image is rejected. Poor-quality images are additionally categorized as "smudged" or "dry". An example of quality estimation using Gabor filters is shown in Fig. 4 for two fingerprints of different quality.

*3) Based on Pixel Intensity:* The method described in [21] classifies blocks into "directional" and "nondirectional" as follows. The sum of intensity differences $D_d(i,j)$ between a pixel $(i,j)$ and $l$ pixels selected along a line segment of direction $d$ centered at $(i,j)$ is computed for $n$ different directions. For each different direction $d$, the histogram of $D_d(i,j)$ values is obtained for all pixels within a given foreground block. If only one of the $n$ histograms has a maximum value that is greater than a prominent threshold, the block is marked as "directional." Otherwise, the block is marked as "nondirectional." An overall quality score $Q$ is finally computed. A relative weight $w_i$ is assigned to each foreground block based on its distance to the centroid of the foreground. The quality score $Q$ is defined as $Q = \sum_D w_i / \sum_F w_i$, where $D$ is the set of directional blocks and $F$ is the set of foreground blocks. If $Q$ is lower than a threshold, then the image is considered to be of poor quality. Measures of the smudginess and dryness of poor-quality images are also defined in [21].

Two methods based on pixel intensity are presented in [11]. The first one measures the variance in gray levels in overlapped blocks. High-quality blocks will have large variance while low-quality blocks will have a small one. The second method measures the local contrast of gray values among ridges and valleys along the local direction of the ridge flow. Blocks with high quality will show high contrast, which means that ridges and valleys are well separated on the grayscale. Shi *et al.* [22] define further features extracted from the gray-level image to characterize a block of a fingerprint image: mean, variation, contrast, and eccentric moment.

The method presented in [23] checks the consistency of ridge and valley's gray level as follows. It first binarizes image blocks using Otsu's method [31] to extract ridge and valley regions and then computes a clustering factor, defined as the degree to which gray values of ridge/valley pixels are clustered. The more clustered the ridge or valley pixels are, the higher the clarity of such a structure and, hence, its quality.

Chen *et al.* [18] proposed a measure which computes the clarity of ridges and valleys. For each block, they extract the amplitude of the sinusoidal-shaped wave that models ridges and valleys along the direction normal to the local ridge direction [24]. A threshold is then used to separate the ridge region and valley region of the block. The gray-level distribution of the segmented ridges and valleys is computed and the overlapping area of the distributions is used as a measure of clarity of ridges and valleys. For ridges/valleys with high clarity, both distributions should have a very small overlapping area. A global clarity score is finally computed by averaging all of the local clarity measures of the image. An example of quality estimation using the local clarity score is shown in Fig. 5 for two fingerprint blocks of different quality.

*4) Based on Power Spectrum:* The method presented in [23] extracts the sinusoidal-shaped wave along the direction normal to the local ridge direction [24] and then computes its discrete Fourier transform. Low-quality blocks will not exhibit an obvious dominant frequency, or it will be out of the normal ridge frequency range.

*5) Based on a Combination of Local Features:* Hong *et al.* [24] modeled ridges and valleys as a sinusoidal-shaped wave along the direction normal to the local ridge direction and extracted the amplitude, frequency, and variance of the sinusoid. Based on these parameters, they classify blocks as recoverable and unrecoverable.

The minutia detection (MINDTCT) package of the NIST Fingerprint Image Software (NFIS) [25] locally analyzes the fingerprint image and generates an image-quality map. The quality of each block is assessed by computing several maps: direction map, low contrast, low flow, and high curve. The direction map is indicating areas of the image with sufficient ridge structure. The low contrast map is marking blocks with weak contrast, which are considered background blocks. The low flow map represents blocks that could not be assigned a dominant ridge flow. The high curve map is marking blocks that are in high curvature areas, which usually are core and delta regions, but also other low-quality regions. These maps are integrated into one quality map, containing five levels of quality (an example is shown in Fig. 6 for two fingerprints of different quality).

*B. Methods Based on Global Features*

Methods that rely on global features analyze the image in a holistic manner and compute a global measure of quality based on the features extracted.

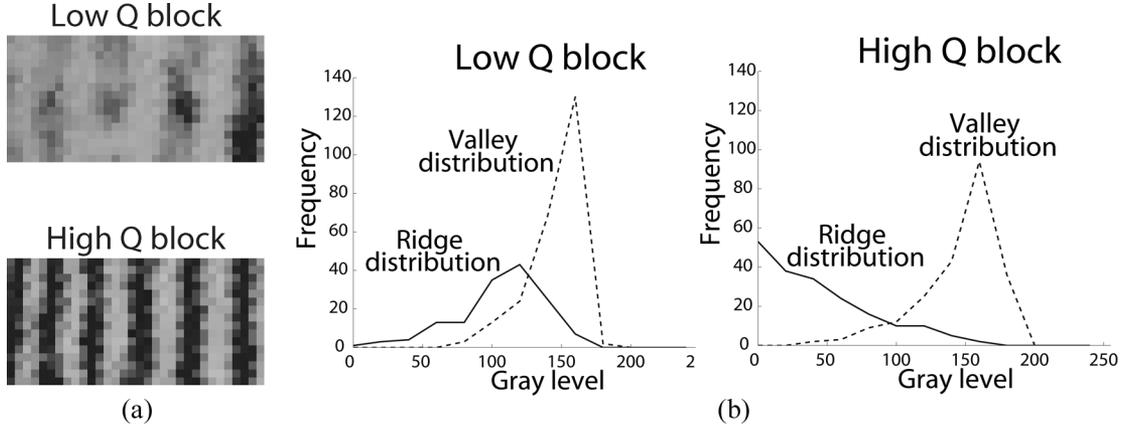

Fig. 5. Computation of the local clarity score for two fingerprint blocks of different quality. Panel (a) is the fingerprint blocks. Panel (b) is the gray-level distributions of the segmented ridges (solid line) and valleys (dashed line). The degree of overlapping for the low- and high-quality block is 0.22 and 0.10, respectively.

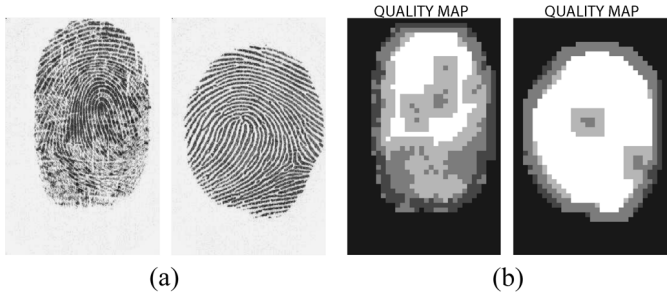

Fig. 6. Fingerprint-quality maps [panel (b)] provided by the minutia detection package of the NIST Fingerprint Image Software for two fingerprints of different quality [panel (a)].

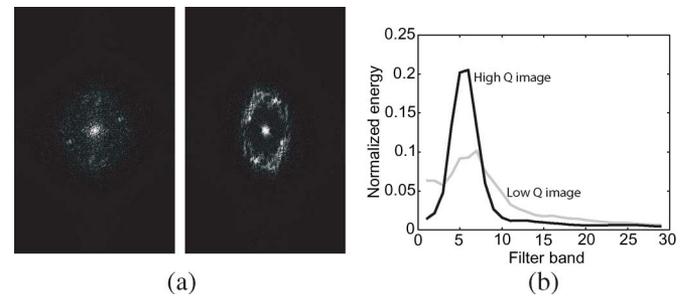

Fig. 7. Computation of the energy concentration in the power spectrum for two fingerprints of different quality. Panel (a) is the power spectra of the images shown in Fig. 1(a). Panel (b) shows the energy distributions in the region of interest. The quality values for the low- and high-quality image are 0.35 and 0.88, respectively.

*1) Based on the Direction Field:* Lim et al. [17] presented two features to analyze the global structure of a fingerprint image. Both of them use the local direction information provided by the direction field, which is estimated in nonoverlapping blocks. The first feature checks the continuity of the direction field. Abrupt direction changes between blocks are accumulated and mapped into a global direction score. As we can observe in Fig. 2, the ridge direction changes smoothly across the whole image in case of high quality. The second feature checks the uniformity of the frequency field [2]. This is accomplished by computing the standard deviation of the ridge-to-valley thickness ratio and mapping it into a global score, as large deviation indicates low image quality.

*2) Based on Power Spectrum:* The global structure is analyzed in [15] by computing the 2-D discrete Fourier transform (DFT). For a fingerprint image, the ridge frequency values lie within a certain range. A region of interest (ROI) of the spectrum is defined as an annular region with a radius ranging between the minimum and maximum typical ridge frequency values. As the fingerprint image quality increases, the energy will be more concentrated within the ROI, see Fig. 7(a). The global quality index $Q_F$ defined in [15] is a measure of the energy concentration in ring-shaped regions of the ROI. For this purpose, a set of bandpass filters is employed to extract the energy in each frequency band. High-quality images will have the energy concentrated in few bands while poor ones will have a more diffused distribution. The energy concentration is measured using the entropy.

An example of quality estimation using the global quality index $Q_F$ is shown in Fig. 7 for two fingerprints of different quality.

### C. Methods Based on Classifiers

The method that uses classifiers [32], [33] defines the quality measure as a degree of separation between the match and nonmatch distributions of a given fingerprint. This can be seen as a prediction of the matcher performance. Tabassi et al. [32], [33] extract the fingerprint features (minutiae in this case) and then compute the quality of each extracted feature to estimate the quality of the fingerprint image, which is defined as stated before.

Let $s(x_{ii})$ be the similarity score of a genuine comparison (match) corresponding to the subject $i$, and $s(x_{ji})$, $i \neq j$ be the similarity score of an impostor comparison (nonmatch) between subject $i$ and impostor $j$. Quality $Q_N$ of a biometric sample $x_{ii}$ is then defined as the prediction of

$$o(x_{ii}) = \frac{s(x_{ii}) - E\left[s(x_{ji})\right]}{\sigma\left(s(x_{ji})\right)} \qquad (1)$$

where $E[.]$ is the mathematical expectation and $\sigma(.)$ is the standard deviation. Equation (1) is a measure of separation between the match and the nonmatch distributions, which is supposed to be higher as image quality increases. The prediction of $o(x_{ii})$ is accomplished by using a neural network. The output of the neural network is a number that classifies the quality of the fingerprint into five values: 5 (poor), 4 (fair), 3 (good), 2 (very good), and 1 (excellent).

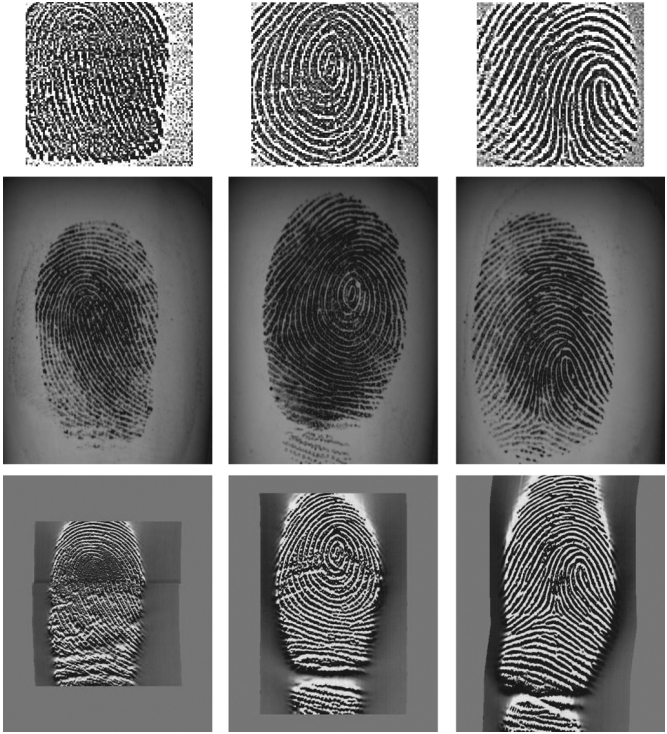

Fig. 8. Example images from the BioSec baseline corpus. Fingerprint images are plotted for the same finger for 1) capacitive sensor (top row), optical sensor (medium row), thermal sensor (bottom row), and 2) three different fingerprints, one per column.

### III. EXPERIMENTS

The aim of our experiments is to compare the behavior of a representative set of quality measures by studying both their correlation and their utility. For the utility study, we compare the impact of the selected image-quality measures in the performance of a widely available fingerprint matcher that uses minutiae to represent and match fingerprints [25]. Minutiae matching is certainly the most well-known and widely used method for fingerprint matching, thanks to its analogy with the way forensic experts compare fingerprints and its acceptance as a proof of identity in the courts of law [2].

#### A. Fingerprint Verification Matcher

As a fingerprint matcher for our study, we use the minutia-based matcher included in the freely available NIST Fingerprint Image Software 2-NFIS2 [25]. For our evaluation and tests with NFIS2, we have used the following packages: 1) MINDTCT for minutia extraction and quality assessment and 2) BOZORTH3 for fingerprint matching. MINDTCT takes a fingerprint image and locates all minutiae in the image (including location, direction, type, and quality). Minutiae extraction is performed in MINDTCT by means of binarization and thinning, as done by most of the proposed extraction methods [2]. The BOZORTH3 matching algorithm computes a matching score between the minutiae templates from two fingerprints. The BOZORTH3 matcher uses only the location and direction of the minutiae points to match the fingerprints, in a translation and rotation–invariant manner. Additional details of these packages can be found in [25].

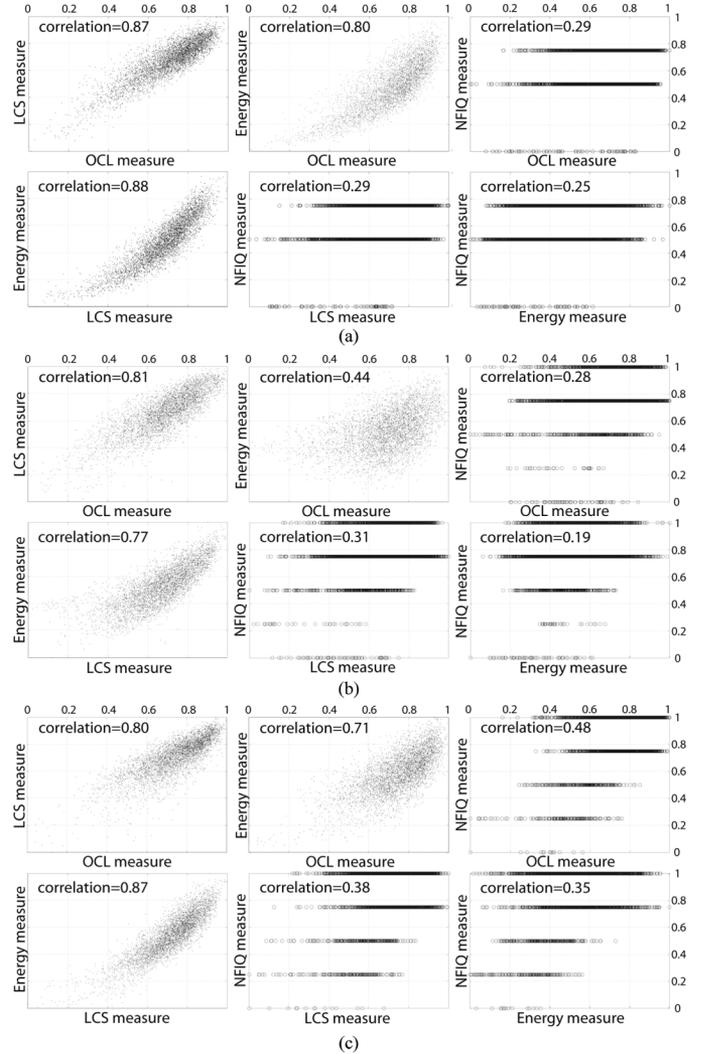

Fig. 9. Correlation between the automatic quality assessment algorithms tested in this work ($x$ and $y$ axis are the quality values of the two algorithms under comparison). The Pearson correlation value between the two algorithms is also shown in each subplot. (a) Capacitive sensor. (b) Optical sensor. (c) Thermal sensor.

#### B. Selected Quality Measures

Different measures have been selected from the literature in order to have a representative set. We have implemented at least one measure that makes use of the different features presented in Tables I and II: direction information (local direction, Gabor filters, or global direction field), pixel intensity information, and power spectrum information. The measure that relies on direction information is the orientation certainty level (OCL) [17], the measure based on pixel intensity information is the local clarity score (LCS) [18], and the measure based on the power spectrum is the energy concentration [15]. We have also used the existing measure based on classifiers, NFIQ, which is included in the NIST Fingerprint Image Software 2-NFIS2 [25].

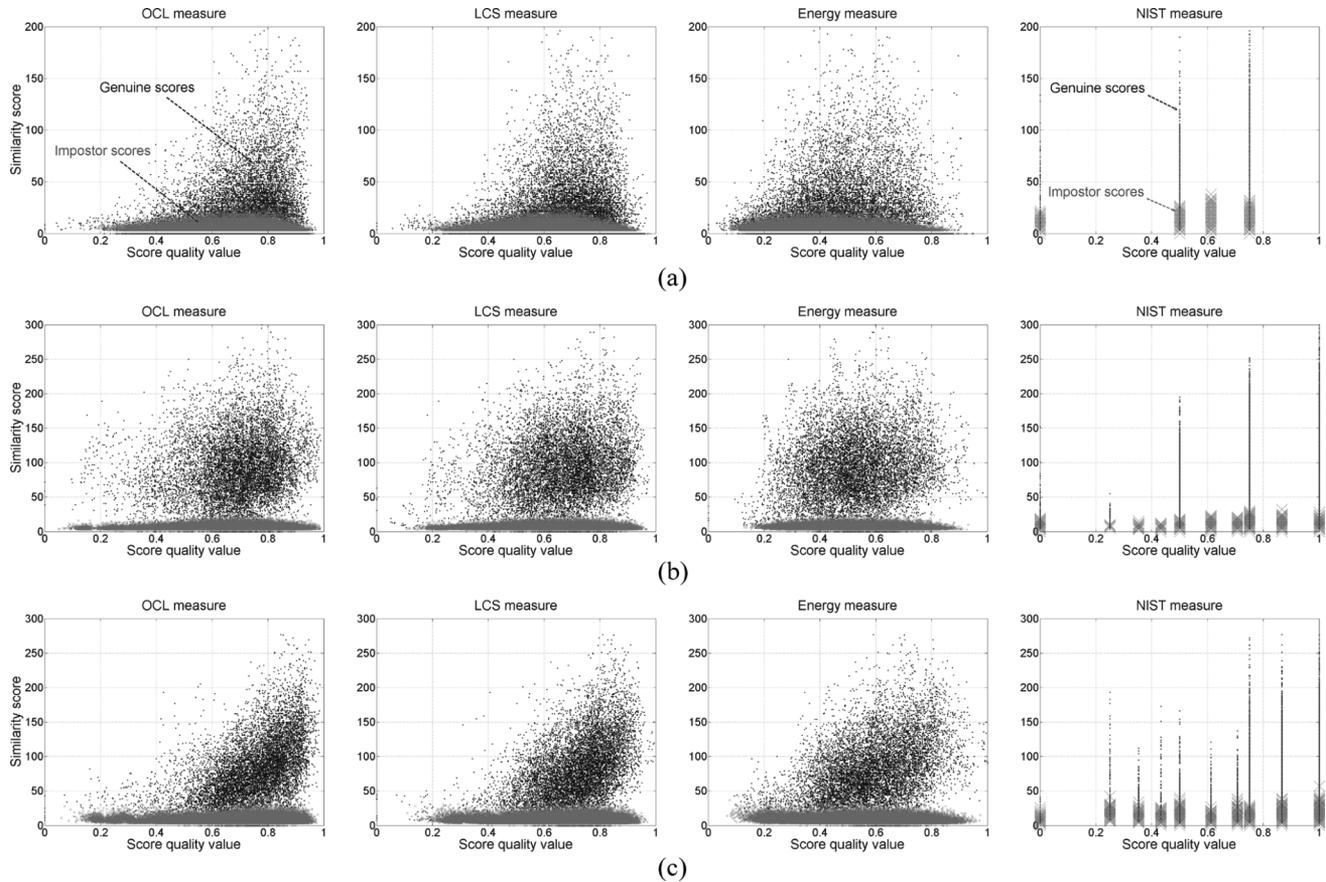

Fig. 10. Dependence of similarity scores ($y$ axis) on the average quality of the template and the input images ($x$ axis). We assign a quality value to a given score, which is computed as $\sqrt{Q_e \times Q_t}$, where $Q_e$ and $Q_t$ are the quality values of the enrolment and test fingerprint samples, respectively, corresponding to the matching. (a) Capacitive sensor. (b) Optical sensor. (c) Thermal sensor.

In the experiments carried out in this paper, all image-quality values are normalized into the [0–1] range, with 0 corresponding to the worst quality and 1 corresponding to the best quality.

### C. Database and Protocol

For the experiments in this paper, we use the BioSec baseline corpus [16]. The data consist of 19 200 fingerprint images acquired from 200 individuals in two acquisition sessions, separated typically by one to four weeks, using three different sensors. The fingerprint sensors are: 1) capacitive sensor Authentec AES400, with an image size of 96 pixels width and 96 pixels height; 2) thermal sensor Atmel FCDEM04, with an image size of 400 pixels width and 496 pixels height; and 3) optical sensor Biometrika FX2000, with an image size of 400 pixels width and 560 pixels height. The capacitive sensor has a resolution of 250 dpi,[2] whereas the thermal and the optical ones have a resolution of 500 dpi. A total of four captures of the print of four fingers (right and left index and middle) were captured with each of the three sensors, interleaving fingers between consecutive acquisitions. The total number of fingerprint images is therefore 200 individuals × 2 sessions × 4 fingers × 4 captures= 6400 images per sensor. In Fig. 8, some fingerprint samples from the BioSec baseline corpus are shown.

The 200 subjects included in BioSec Baseline are further divided into: 1) the development set, including the first 25 and the last 25 individuals of the corpus, totaling 50 individuals and 2) the test set, including the remaining 150 individuals. The development set is used to tune the parameters of the different quality assessment algorithms. No training of parameters is done on the test set. We consider the different fingers of the test set as different users enrolled in the system, thus resulting in 150 × 4 = 600 users. For evaluation of the verification performance, the following matchings are defined in the test set: 1) genuine matchings: the four samples in the first session to the four samples in the second session, resulting in 150 individuals × 4 fingers × 4 templates × 4 test images= 9600 genuine scores per sensor and 2) impostor matchings: the first sample in the first session to the same sample of the remaining users, avoiding symmetric matches, resulting in (150 × 4) × (150 × 4 − 1)/2= 179 700 impostor scores per sensor.

### D. Results and Discussion

In Fig. 9, we can observe the correlation among the quality measures tested in this paper. In addition, the Neyman–Pearson correlation values [34] between the measures are shown. We observe high correlation between all measures, except when the NFIQ one is involved. This could be because of the finite number of quality labels used by this algorithm [25]. It is worth noting that the lowest correlation values are obtained with the optical sensor.

---
[2]The NIST-NFIQ quality measure is developed for 500-dpi images, thus images from the capacitive sensor are first interpolated using bicubic interpolation.

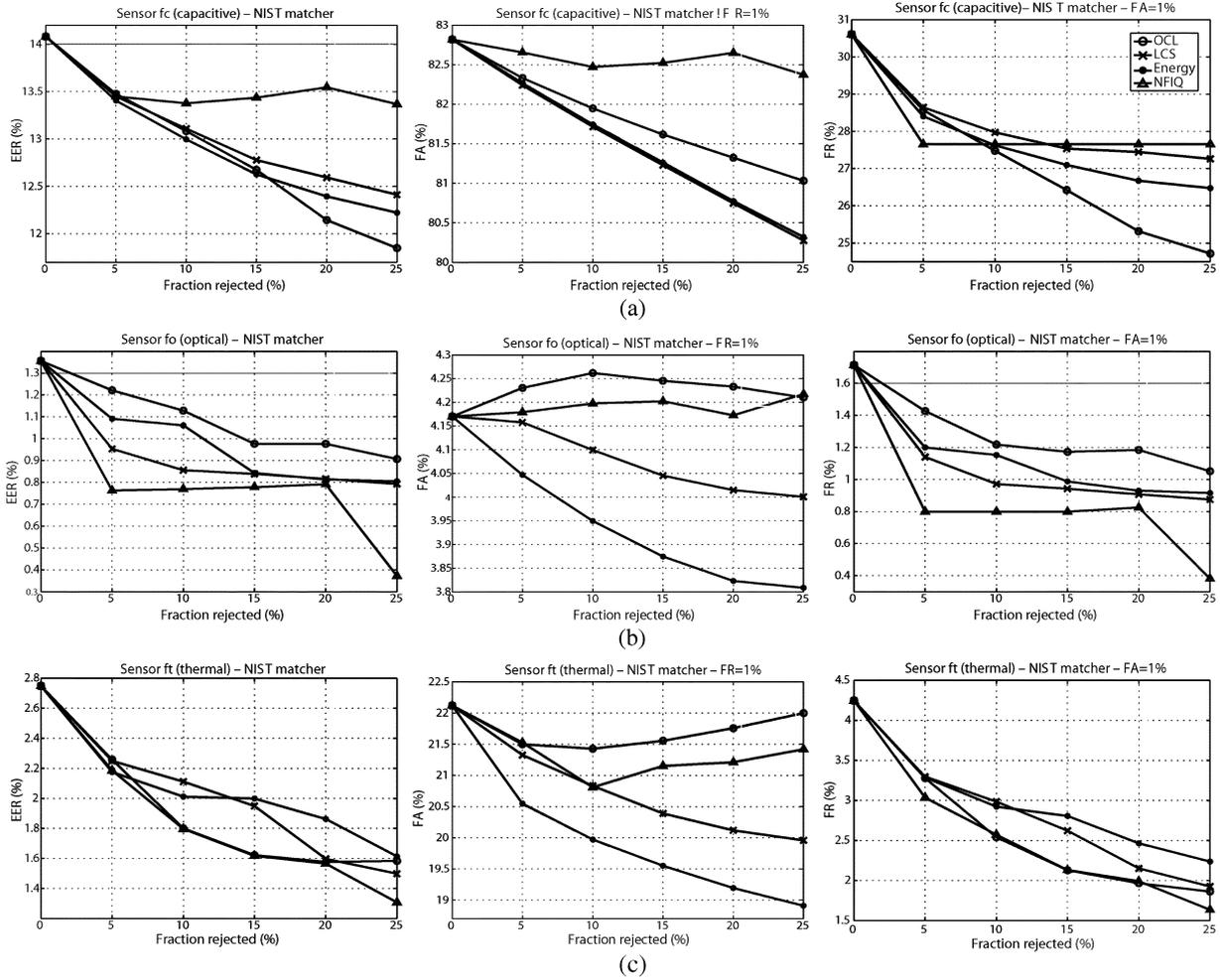

Fig. 11. Verification performance as samples with the lowest quality value are rejected. Results are shown for all of the quality measures tested in this work in terms of EER (first column). The false acceptance rate at 1% FRR (second column) and false rejection rate at 1% FAR (third column). (a) Capacitive sensor. (b) Optical sensor. (c) Thermal sensor.

In order to evaluate the utility of the compared quality metrics (i.e., their capacity to predict the performance [14]), we plot in Fig. 10 the similarity scores against the average quality of the two involved fingerprint images. We assign a quality value to a given score, which is computed as $\sqrt{Q_e \times Q_t}$, where $Q_e$ and $Q_t$ are the quality values of the enrolment and test fingerprint, respectively, corresponding to the matching (note that the NIST–NFIQ quality measure only provides five possible values for $Q_e$ and $Q_t$ and, thus, the combined value $\sqrt{Q_e \times Q_t}$ also exhibit a discrete nature but with more than five possible values). We observe some degree of correlation between the genuine similarity scores and the quality values in Fig. 10. On the other hand, almost no correlation is observed between quality and impostor scores, as can be seen in Fig. 10 for most cases. We also observe a desirable fact in Fig. 10: in low-quality conditions, impostor scores should remain low.

Fig. 11 depicts the error rates of the verification system as we reject samples (i.e., matching scores) with the lowest quality value. We observe that, in general, the performance metrics improve when samples with the lowest quality are rejected (e.g., a decrease of either the false acceptance rate (at 1% FRR), the false rejection rate (at 1% FAR), or the equal error rate is observed). Since there is high correlation between genuine scores and quality, the best improvement is obtained in the FRR. After rejection of just 5% of the samples, FRR is improved in the best case about 10%, 50%, and 30% for the capacitive, optical, and thermal sensor, respectively. Significative improvement is also obtained in the EER (3.5%, 45%, and 21%, respectively, in the best case). The lowest improvement is obtained with the capacitive sensor, as a consequence of its smaller acquisition surface and lower resolution (see Fig. 8). For a quality algorithm to be effective, improvements in the FAR when quality increases are also expected [4]. However, as there is almost no correlation between the quality and impostor scores (see Fig. 10), smaller improvement is obtained for the FAR (2.73% and 6.8% for the optical and thermal sensor, respectively, in the best case), or even no improvement, as observed in some cases.

It is worth noting that similar performance variations are observed in Fig. 11 for the capacitive and thermal sensors with most of the quality algorithms. This is not true for the optical sensor, which also showed the lowest correlation values between quality measures.

## IV. CONCLUSION

This paper reviews existing approaches for fingerprint image-quality estimation, including visual examples showing the behavior under different quality conditions. Existing approaches have been divided into: 1) those that use local features of the image; 2) those that use global features of the image; and 3) those that address the problem of quality assessment as a classification problem. Local and global image features are extracted using different sources: direction field, Gabor filter responses, power spectrum, and pixel intensity values.

Previous studies demonstrate that the performance of a fingerprint recognition system is heavily affected by the quality of fingerprint images [6], [7]. In this paper, we study the effect of rejecting low-quality samples using a selection of quality estimation algorithms that includes approaches based on the three classes defined before. We also compare the behavior of the selected quality methods by reporting the correlation between them and the relationship between quality and similarity scores. We use for our experiments the BioSec multimodal baseline corpus, which includes 19 200 fingerprint images from 200 individuals acquired in two acquisition sessions using a capacitive, an optical, and a thermal fingerprint sensor. Experimental results show high correlation between genuine scores and quality, whereas almost no correlation is found between impostor scores and the quality measures. As a result, the highest improvement when rejecting low-quality samples is obtained for the false rejection rate at a given false acceptance rate.

High correlation is found between quality measures in most cases. However, different correlation values are obtained depending on the sensor. This suggests that quality measures work differently with each sensor, which will be a source of future work. Due to their different physical principles, some quality measures could not be suitable for a certain kind of sensor. On the other hand, different quality measures could provide complementary information, and its combination may improve the process of assessing the quality of a fingerprint image. Lastly, future work also includes the study of the effects of low-quality images in systems that use alternative methods for minutiae extraction [35], [36] or alternative features for fingerprint matching (e.g., ridge information [37] or gray local information [38]).

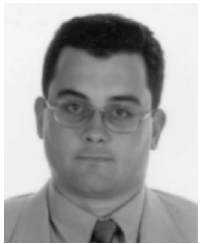

**Fernando Alonso-Fernandez** (S'04) received the M.S. degree in electrical engineering from the Universidad Politecnica de Madrid, Madrid, Spain, in 2003, where he is currently pursuing the Ph.D. degree in biometrics.

His research interests include signal and image processing, pattern recognition, and biometrics. He has published several journal and conference papers and is actively involved in European projects focused on biometrics (e.g., BioSec IP and Biosecure NoE).

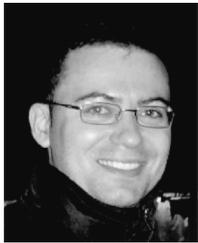

**Julian Fierrez** (M'06) received the M.Sc. and the Ph.D. degrees in electrical engineering from the Universidad Politecnica de Madrid, Madrid, Spain, in 2001 and 2006, respectively.

Currently, he is an Assistant Researcher with the Universidad Autonoma de Madrid, Madrid, Spain. He is a Visiting Researcher with Michigan State University, East Lansing. His research interests include signal and image processing, pattern recognition, and biometrics.

Dr. Fierrez is actively involved in European projects focused on biometrics (e.g., Biosecure NoE) and is the recipient of a number of distinctions, including: Best Poster Award at AVBPA 2003, 2nd best signature verification system at SVC 2004, Rosina Ribalta Award to the best Spanish Ph.D. proposal in 2005, Motorola best student paper at ICB 2006, and EBF European Biometric Industry Award 2006.

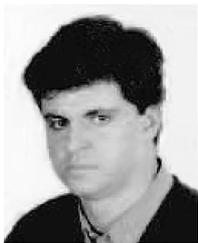

**Javier Ortega-Garcia** (M'96) received the M.Sc. degree in electrical engineering (Ing.Tel.) and the Ph.D. degree (Hons.) in electrical engineering (D.Ing.Tel.) from the Universidad Politécnica de Madrid, Madrid, Spain, in 1989 and 1996, respectively.

Dr. Ortega-Garcia is Founder and Co-Director of the ATVS/Biometric Recognition Group. Currently, he is an Associate Professor with the Escuela Politécnica Superior, Universidad Autónoma de Madrid, Madrid, Spain, where he teaches digital signal-processing and speech-processing courses. He also teaches a Ph.D. degree course in biometric signal processing. His research interests are biometrics signal processing, speaker recognition, face recognition, fingerprint recognition, online signature verification, data fusion, and multimodality in biometrics. He has been published in many international contributions, including book chapters, refereed journal, and conference papers.

Dr. Ortega-Garcia chaired "Odyssey-04, The Speaker Recognition Workshop", cosponsored by the IEEE.

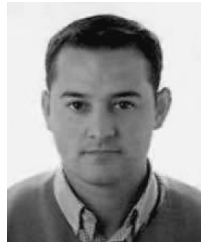

**Joaquin Gonzalez-Rodriguez** (M'96) is an Associate Professor at the Universidad Autonoma de Madrid, Madrid, Spain. He is Co-Leader of an active research group that focuses on speech processing (speaker and language recognition, forensic applications, speech enhancement, array processing) and biometrics (speaker, fingerprint, face, and online signature recognition). He has led European and national research projects, such as secure access front end (SAFE), EU-FP5 trial startup, participates in an IP and an NoE in the EU-FP6, and is a Scientific Reviewer in different international conferences and journals (IEEE TRANSACTIONS ON SPEECH AND SIGNAL PROCESSING). He has led ATVS/Biometric Recognition Group submissions to the National Institute of Standards Technology (NIST) 2001/02/04/05 Speaker Recognition Evaluations and to the National Institute of Standards Technology 2005 Language Recognition Evaluation.

Prof. Gonzalez-Rodriguez is an active member of the Forensic Speech community, contributing to ENFSI and IAFP meetings. He was Vice Chairman of Odyssey 2004, the Speaker Recognition Workshop that was hosted by his group in 2004. He is also member of the IEEE Signal Processing Society and International Speech Communication Association (ISCA).

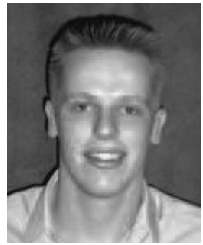

**Hartwig Fronthaler** received the M.Sc. degree from Halmstad University, Halmstad, Sweden.

His specialization has been in the field of image analysis with a focus on biometrics. He joined the signal analysis group of Halmstad University in 2004. His research interests are in the field of fingerprint processing, including automatic quality assessment and feature extraction. He has been involved in research on face biometrics.

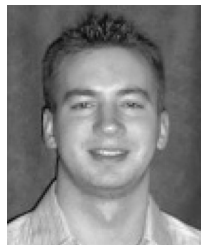

**Klaus Kollreider** received the M.S. degree in computer systems engineering from Halmstad University, Halmstad, Sweden, in 2004.

His research interests include signal analysis and computer vision, in particular, face biometrics and antispoofing measures by object detection and tracking. He is involved in a European project focused on biometrics BioSecure, where he has also contributed to a reference system for fingerprint matching.

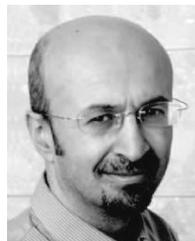

**Josef Bigun** (F'03) received the M.S. and Ph.D. degrees from Linkoeping University, Linkoeping, Sweden, in 1983 and 1988, respectively.

From 1988 to 1998, he was with the Swiss Federal Institute of Technology in Lausanne (EPFL), Lausanne, Switzerland. His research interests include a broad field in computer vision, texture and motion analysis, biometrics, and the understanding of biological-recognition mechanisms.

Dr. Bigun was elected Professor to the Signal Analysis Chair at Halmstad University and Chalmers University of Technology, Gothenburg, Sweden, in 1998. He has co-chaired several international conferences and has contributed as a referee or as an editorial board member of journals including PRL and IEEE TRANSACTIONS ON IMAGE PROCESSING. He served in the executive committees of several associations including IAPR. He has been elected Fellow of IAPR.